\documentclass[letterpaper]{article}

\usepackage{natbib,alifeconf}  %% The order is important

\usepackage[final]{dkadish}

\usepackage[english]{babel}
\usepackage[autostyle, english = american]{csquotes}
\MakeOuterQuote{"}

\usepackage[utf8]{inputenc}
\usepackage[T1]{fontenc}
\usepackage{csquotes}

\usepackage[colorinlistoftodos,prependcaption,textsize=tiny]{todonotes}

\usepackage{pdfpages}
\usepackage{graphicx}
\usepackage[font=small]{subcaption}

\usepackage{lipsum}
\usepackage{booktabs}

\usepackage{mathtools, nccmath}

\usepackage{glossaries}
\makeglossaries

\newglossaryentry{ecology}
{
  name=ecology,
  description={DEF} 
}

\newglossaryentry{niche}
{
  name=niche,
  description={DEF} 
}

\newglossaryentry{fundamental niche}
{
  name=fundamental niche,
  description={the range of possibilities for a niche}
}

\newglossaryentry{niche construction}
{
  name=niche construction,
  description={DEF} 
}

\newglossaryentry{niche differentiation}
{
  name=niche differentiation,
  description={DEF} 
}

\newglossaryentry{competitive exclusion}
{
  name=competitive exclusion,
  description={DEF} 
}

\newglossaryentry{polyculture}
{
  name=polyculture,
  description={an ecosystem in which many different species of plant are raised together, as opposed to a monoculture where many different things are grown together.}
}

\newglossaryentry{bid}
{
  name={bio-inspired design},
  description={DEF} 
}

\newglossaryentry{eid}
{
  name={ecology-informed design},
  description={DEF} 
}

\newglossaryentry{add}
{
  name=design-by-analogy,
  description={DEF} 
}

\newacronym[]{voc}{VOC}{volatile organic compound}

\newacronym[]{mos}{MOS}{metal oxide}

\newacronym[]{anh}{ANH}{acoustic niche hypothesis}

\newacronym[]{sac}{SAC}{synthetic acoustic ecology}

\newacronym[]{alife}{ALife}{artificial life}

\newacronym[]{neat}{NEAT}{neuroevolution of augmenting topologies}

\newacronym[]{tsne}{t-SNE}{t-distributed Stochastic Neighbour Embedding}

\begin{document}

\newcommand{\ttl}[0]{An \acrlong{alife} approach to studying niche differentiation in soundscape ecology}

\newcommand{\spec}[1]{{\color{ForestGreen}#1}}
\newcommand{\togen}[1]{{\color{Maroon}#1}}

\title{\ttl{}}%\\ \normalsize{\sttl}}
\author{David Kadish \and Sebastian Risi \and Laura Beloff
\mbox{}\\
IT University of Copenhagen, Copenhagen, Denmark \\
davk@itu.dk} % email of corresponding author
%\date{March 8, 2019}%\today}
\maketitle

%Paper and Abstract: March 8, 2019\\
%Full papers have an 8-page maximum length and should report on new, unpublished work.
%
%\input{sections/outline}

%\input{sections/notes.tex}

\begin{abstract}
	\Acrlong{alife} simulations are an important tool in the study of ecological phenomena that can be difficult to examine directly in natural environments.
	Recent work has established the soundscape as an ecologically important resource and it has been proposed that the differentiation of animal vocalizations within a soundscape is driven by the imperative of intraspecies communication.
	The experiments in this paper test that hypothesis in a simulated soundscape in order to verify the feasibility of intraspecies communication as a driver of acoustic niche differentiation.
	The impact of intraspecies communication is found to be a significant factor in the division of a soundscape's frequency spectrum when compared to simulations where the need to identify signals from conspecifics does not drive the evolution of signalling.
	The method of simulating the effects of interspecies interactions on the soundscape is positioned as a tool for developing \acrlong{alife} agents that can inhabit and interact with physical ecosystems and soundscapes.{\let\thefootnote\relax\footnote{{The manuscript is slightly amended from the published version to correct an error in \autoref{fig:cluster}. The figure in the published version plots the messages from a different run of the simulation than the one shown in \autoref{fig:silhouette}. This in no way changes the results of the study.}}}
\end{abstract}

\section{Introduction}

\Acrlong{alife} experiments have become important tools for exploring biological phenomena.
In particular, they have allowed researchers to study the relationships between evolutionary processes and ecological theories~\citep{Aguilar2014}, like the emergence of interspecies relationships like mutualism and parasitism~\citep{Watson2000}.

One area of ecology that has received little attention thus far from \gls{alife} studies is soundscape ecology.
The field of soundscape ecology has been formalized by researchers over the past decade~\citep{Pijanowski2011a}, building on earlier conceptions of the soundscape~\citep{Schafer1977a}.
One of its foundational theories is the \gls{anh}~\citep{Krause1987}, which applies the concept of ecological niches --- the distribution of resources that are used by a species in an ecosystem~\citep{Pocheville2015} --- to the soundscape.

%TODO RESP: "It would perhaps have been interesting to allow more than one way of variation. For example, it might be interesting to see what would have happened while allowing agents to vary contemporarily the frequency and the message semantics (a semantics that should emerge from the interaction among senders and receivers).  But, again, the risk of circular reasoning would be similar." discuss how the result is expected but still interesting

This experiment tests the proposed mechanisms for the formation of these niches in a virtual soundscape in order to understand how species change vocalizations in response to one another.
It models the behaviour of two species in a virtual ecosystem and tracks how their calls shift through the audio spectrum in response to different evolutionary pressures.
Through the experiment, evolutionary pressure to communicate within a species is found to play a significant role in the formation of acoustic niches.
%TODO RESP: R2 "discuss what kind of novel findings were obtained from the results, comparing those from previous related models" || ODO SB: Replace with “The main results of this study...” Perhaps put before this...Put the main contribution in the intro here. ______ Point out in introduction: we understand that this is simplified, but it allows exploration

In examining the emergence of communication between artificially evolved species, this study draws from a body~\citep{Arita1998,Wagner2000,Sasahara2007} of \gls{alife}-based studies of communication including the work of \citet{Floreano2007a} in emergent communication between robotic agents.
However, it is distinct from these previous studies in its focus on the effect of the emergent communication on the ecological phenomena of niche differentiation.

The main contributions of this study are the development of a simplified model of a soundscape for the purpose of rapid experimentation and in-depth analysis of population-soundscape dynamics, and the demonstration of the \gls{anh} on this model.

\section{Background}

%TODO RESP || SB: Mention Dario & Sarah Mitri, evolution of communication. Yes, we’re evolving communication, but what we’re looking at is how that leads to differentiated niches. While others have looked at the evolution of communication, we look at

In the physical world, the concept of soundscape --- the collection of the acoustic features of a landscape --- has roots and influences in a diverse array of academic fields~\citep{Lyonblum2017}.
It grew initially out of the arts and cultural studies work of \citet{Westerkamp1974}, \citet{Schafer1977a}, and \citet{Truax1978}, but has since expanded into the sciences.
In the field of ecology, the soundscape is considered an important ecological resource and its composition is thought to indicate the diversity and stability of the ecosystem~\citep{Pijanowski2011}.
Though the field of soundscape ecology was only proposed relatively recently~\citep{Pijanowski2011}, the application of ecological principles to the study of soundscape has a longer history.
Notably, the concept of ecological niches was first introduced in the context of sonic resources by \citet{Krause1987} as the \acrlong{anh}.

\subsection{\Acrfull{anh}}
The \acrlong{anh} expands the concept of ecological niches to the spectro-temporal plane of the soundscape.
\citeauthor{Krause1987} proposed that, in the same way that niche differentiation leads to species making use of the range of physical resources available in an ecosystem, species tend to differentiate their use of an ecosystem's sonic resources.
This differentiation, according to \citeauthor{Krause1987}, occurs spectrally in the sonic frequencies that animals use for vocalization and temporally in the time-based patterns of their sounds.
The theory holds that older, more mature ecosystems should show a greater degree of differentiation between the auditory niches that long-established species occupy.

%TODO Check these references more closely
The \gls{anh} describes the result of acoustic differentiation, but \citet{Endler1992} proposed the primary mechanism for this evolutionary driver: sexual selection based on a mate's ability to hear a call and the ability to maintain territory.
In this formulation, vocalizations and auditory receptors have co-evolved to maximize the reception of signals from members of ones own species (conspecifics), while minimizing interference from members of other species (heterospecifics).

This type of spectral differentiation has been observed numerous times in the wild: in the calls of certain species of frogs~\citep{Feng2007}; in the buzzing of cicadas~\citep{Sueur2002}; and in the overall division of a soundscape in Borneo between a series of birds, gibbons, and accompanying insects~\citep{Krause2008}.
However, it has proved difficult to experimentally probe the formation and division of spectral niches, due to the lengthy timescales that would be required to allow evolutionary processes to progress~\citep{Miller1995} and the complexity of the systems and soundscapes that are encountered "in the wild"~\citep{Wheeler2002}.

\subsection{\Gls{alife} approaches}
Where ecological phenomena have been difficult to experimentally investigate, researchers have proposed that \gls{alife} approaches can be a mode of inquiry that allows for the manipulation of particular conditions and the rapid collection of large quantities of data about a simulated ecological system~\citep{Miller1995}.
In \citeyear{Eldridge2018}, \citeauthor*{Eldridge2018} proposed \gls{sac} as a toolset for exploring questions in the field of soundscape ecology using \gls{alife} methods in virtual ecosystems.
Their study examined one of the assertions of \gls{anh}~\citep{Krause1987} --- that one can identify the maturity of an ecosystem by examining its acoustic signature.
Using a multi-agent system model, they demonstrated that patterns emerge in two common acoustic indices that indicate the stability of a model ecosystem.

\subsection{Niche differentiation mechanisms}
The study in this paper uses a virtual soundscape to test hypotheses in soundscape ecology, building on the work of \citet{Eldridge2018}.
While \citeauthor{Eldridge2018}'s study focused on the verification of acoustic biodiversity metrics, this study examines the mechanisms that breed interspecific diversity and intraspecific convergence in the vocalizations of communities in a soundscape.
In particular, it is designed to test \citeauthor{Endler1992}'s hypothesis~\citeyearpar{Endler1992} that the ability to identify vocalizations from members of the same species drives acoustic niche differentiation.

The \acrlong{anh} posits that soundscapes niches are differentiated on both spectral and temporal levels, so that species ensure that their calls are isolated in both frequency and time.
In order to simplify the modelling and analysis and to allow for a deeper examination of the effects of differentiation, this study focuses only on the spectral component of this differentiation.

\newcommand{\lbl}[1]{(\textit{#1})}

\section{Approach}

\begin{figure*}
	\centering
	\includegraphics[width=\linewidth]{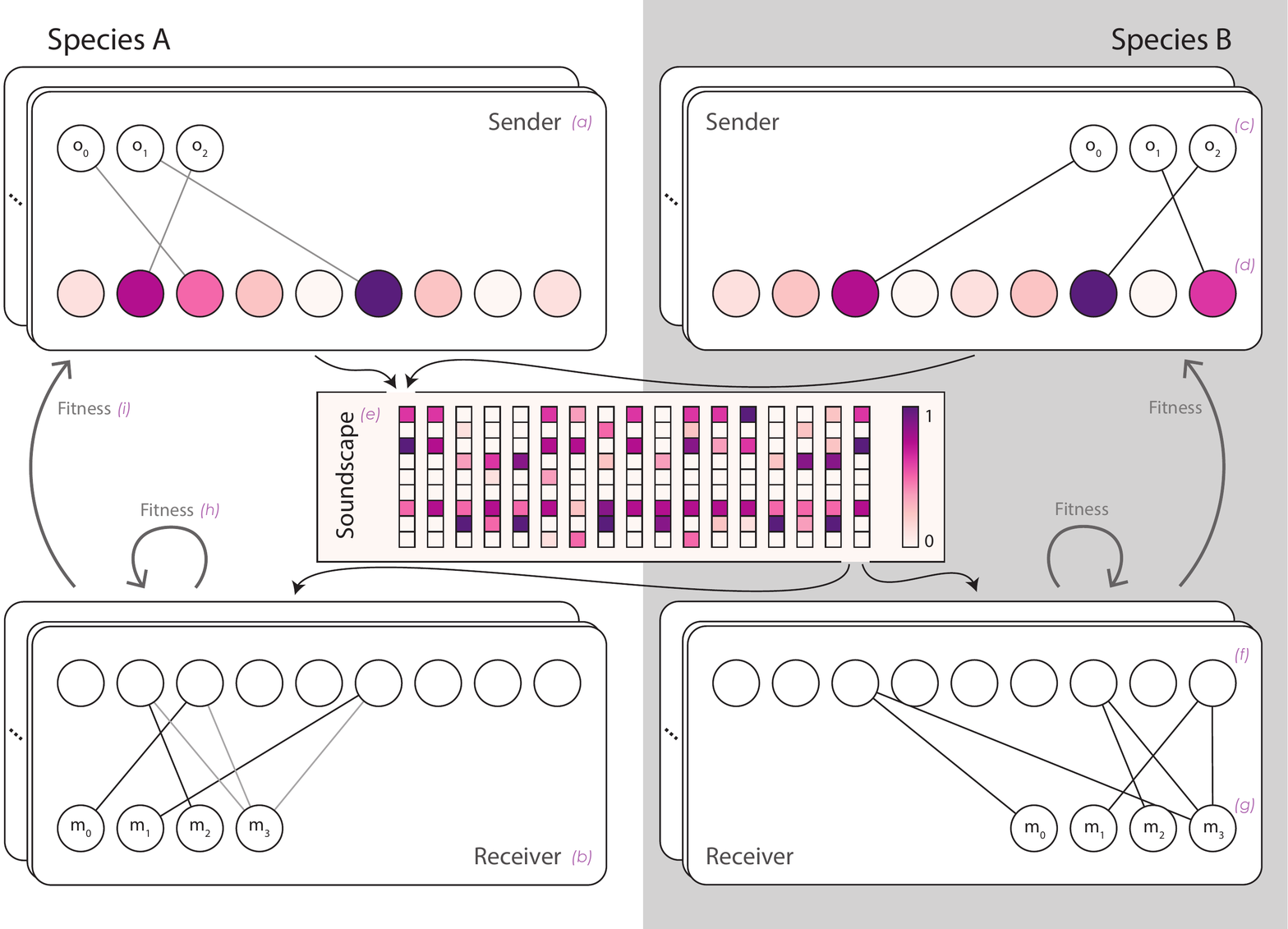}
	\caption[The experimental setup]{The experimental setup. Senders \lbl{a} encode a 3-bit message ($o_{0..2}$) into the 9-band soundscape \lbl{e} using a neural network with 3 inputs \lbl{c} and 9 outputs \lbl{d}. Receivers \lbl{b} "hear" encoded messages from all species' senders and predict the original message ($m_{0..2}$) and whether the message originates from a conspecific ($m_3$) using the 9 inputs \lbl{f} and 4 outputs \lbl{g} of their neural networks. Sender fitness \lbl{i} depends on how well conspecific receivers identify their species and decode their messages. Receiver fitness \lbl{h} depends on how well they identify the species of all senders and how well they decode messages from conspecifics.}
	\label{fig:experimental-setup}
\end{figure*}

The experimental setup for testing the drivers of acoustic niche differentiation consists of a set of evolving populations and a soundscape that they communicate within.
The experiment tests two hypotheses: the alternative hypothesis ($H_1$), that acoustic niche spectral differentiation is driven a need to identify signals from potential mates or territorial rivals of the same species; and the null hypothesis ($H_0$) that spectral differentiation in acoustic niches is not driven by the need to distinguish the species of the signaller.

In order to facilitate rapid experimentation and ease the analysis of the emergent signalling systems, the experiments use a simplified, discretized model of a soundscape instead of a full-spectrum, temporally-varying acoustic space.
Sounds are modelled as 9-bit vectors that represent the use of 9 available frequency bands in an instantaneous signal.
These simplifications allow the repetition of the experiments many times with a large number of generations and individuals, such that results reflect general trends in the dynamics of these systems rather than the peculiarities of any single simulation.
%TODO RESP: SB: Simpler model = This system simplifies the analysis. The insights we got (simpler model, faster run) are different to theirs/support what they did? BE SPECIFIC. In this context the simplified model facilities the analysis of the phenomenon.
The entire system is illustrated in \autoref{fig:experimental-setup} and described in detail in the sections below.
Lettering in brackets refers diagram labels in \autoref{fig:experimental-setup}.

\subsection{Populations}

In soundscape ecology in the physical world, the communicative process is often assessed in two parts: sender and receiver.
Every individual, of course, is both sender and receiver, but the processes experience different evolutionary pressures;
"[n]atural selection favors signals that elicit a response in the receiver that increases or maintains the fitness of the sender"~\citep{Endler1992}.
The same is true in reverse, such that the sender and receiver of a particular species evolve alongside one another, but with sightly different driving forces.

The populations in this experiment are modelled as artificial neural networks, which are optimized with the \gls{neat} algorithm~\citep{Stanley2002a}.
\gls{neat} models individual phenotypes as neural networks with a fixed number of inputs and outputs and an evolvable internal structure and connectivity.
This allows the population to begin with simple neural structures and to evolve complexity as necessary to achieve the task.

For the experiment presented here, each species actually consists of two \gls{neat} populations: a population of senders (a in \autoref{fig:experimental-setup}) and a population of receivers (b in \autoref{fig:experimental-setup}).
%TODO RESP: DK: Discuss as INTERPRETATION as frequency bands...Why? Allows for deeper analysis. SB: Even though it’s an abstract model, you can still gain insights. Note that soundscapes are composed of waveforms, etc. but abstract model can give insights.
The populations communicate over a simulated soundscape (e) that consists of 9-bit vectors, interpreted as acoustic frequency bands which can be used by senders to transmit messages.
The senders encode 3-bit messages ($[o_0, o_1, o_2]$) into a representation in the 9 frequency bands using their 3-input (c), 9-output (d) neural network structures.
The structure of 2 species encoding 3-bit messages into a 9-band soundscape allows for the development of relatively complex messaging while allowing the soundscape to remain undersaturated as each species could theoretically communicate in only 3 of the 9 bands.
The frequency bands form the inputs to the 9-input (f), 4-output (g) receiver neural networks.
The first 3 outputs ($[m_0,m_1,m_2]$) of the receiver network are its estimation of the original message and the final output represents the receiver's prediction of whether the message comes from a conspecific ($m_3 \geq 0.5$) or from a heterospecific individual ($m_3 < 0.5$).

The soundscape (e) is shared among species but messages are received serially in order to decouple timing effects; therefore, each receiver "hears" messages from the senders of all of the present species, but receives them one at a time.
Additionally, any spatial arrangement of the individuals is not considered as part of this experiment, so each receiver "hears" the signals from every sender at the same "volume" with no attenuation due to a distance or set of obstacles between them.

\subsection{Fitness}

In a communicative process, the evolutionary pressure on senders and receivers is related but differs in some crucial aspects.
The fitness functions used in this experiment reflect these differences.
Since communication for mating and territorial maintenance is hypothesized to drive acoustic differentiation~\citep{Endler1992}, the sender is indifferent to how its messages are interpreted by receivers from other species.
The receiver, however, processes all messages regardless of their origin; it has to learn how to differentiate messages from conspecifics from those of heterospecifics.

Following this reasoning, the fitness of the sender (i) is formulated to reflect how well its message is understood --- or correctly decoded --- by the receivers of its own species; it does not depend on how the receivers of the another species process its messages.
The fitness of the receiver (h) reflects both how well it is able to distinguish the species of the sender \textit{as well} as whether it is able to correctly decode the message.

The ability of a receiver to perform these two tasks --- identifying messages from conspecifics and decoding messages --- is formulated into components of the the fitness function as $f_s$ (species identification fitness, \autoref{eq:spec_fit}) and $f_d$ (message decoding fitness, \autoref{eq:bits_fit}).
%TODO LATER This shouldn't be m. Should be s and r for sender and receiver
$m$ is the decoded message where the first three components ($m_{0..2}$) are message as decoded by a receiver.
The fourth value output by the receiver ($m_3$) determines whether the receiver has identified this message as coming from a conspecific ($m_3 > 0.5$) or from a member of another species.
The original message is a three-bit string represented by $o_i$.

\begin{equation} \label{eq:spec_fit}
 f_s(m) = \begin{cases} 
   f_{adj}(1 - \left|1 - m_3\right|) & \text{if } \text{same species} \\
   f_{adj}(1 - \left|0 - m_3\right|) & \text{if } \text{different species}
  \end{cases}
\end{equation}

\begin{equation} \label{eq:bits_fit} 
 f_d(m) = 3 * \prod_{i=0}^{2} f_{adj}\left(1 - \left|o_i - m_i\right|\right)
\end{equation}

To achieve the desired fitness formulations, these equations are applied in different ways for senders and receivers by adjusting the enabling/disabling coefficients $e_s$ and $e_d$ in \autoref{eq:total}.
For each message produced, a sender's fitness is based on the interpretation of the message by all receivers \textit{from its own species}.
\autoref{eq:total} is applied for each receiver from the sender's species with $e_s=1$.
The value of $e_d$ depends on whether the species is identified incorrectly ($e_d=0$) or correctly ($e_d=1$).
IfR the species is incorrectly identified, then the interpretation of the message is of no consequence, which is why the fitness of the message decoding is ignored.

Receivers "hear" messages from the senders from both species and their ability to identify and ignore messages that are not from their species is an important component of their fitness.
For each message that a receiver "hears", $f_s$ is calculated as part of its fitness ($e_s=1$).
If the receiver correctly identifies that a message originated from a member of its own species, it receives an additional score for decoding the bits of the original message ($f_d$) and a bonus multiplier ($f_b$) for correctly identifying multiple bits ($e_d=1$), as described in \autoref{eq:total}.

\begin{equation} \label{eq:bonus}
 f_b(e_s, N) = \begin{cases} 
   1.0 & \text{if } e_s = 0 \\
   \displaystyle \prod_{i=0}^N (\frac{i}{10} + 1) & \text{if } e_s = 1 \\
  \end{cases}
\end{equation}

\begin{equation} \label{eq:total}
f_t = \left(e_s f_s(m) + e_d f_d(m)\right) * f_b(e_s, N)
\end{equation}

One detail that requires some explanation is the adjustment function ($f_{adj}$) applied to the fitness equations for species identification ($f_s$) and message decoding ($f_d$).
The results that these equations evaluate are treated as binary in the operation of the system but the receivers produce output as decimal numbers between 0 and 1.
If the receiver outputs $m_3=0.6$ for a message from a member of its own species, the consequence is no different from $m_3=1.0$ --- the receiver has correctly decided that the message should not be ignored.
However, an application of \autoref{eq:spec_fit} without $f_{adj}$ would result in quite different fitnesses for the two outputs.
\autoref{eq:nonlin} creates a sharp rise in the fitness, centred around a value of $0.5$ without producing a discontinuity, which was found to create an effective fitness landscape for the evolutionary process.

\begin{equation} \label{eq:nonlin}
 f_{adj}(x)=\frac{1}{2} \left(\tanh(8.0 * (x - 0.5)) + 1\right)
\end{equation}

\subsection{Null model and hypothesis}

%TODO RESP: R2 "not clear to me if H0 and the null model is the same" Yes, the null model assumes that discrimination does not affect the fitness WHICH LEADS to differences in how niches are differentiated. || Changed section heading to Null model and hypothesis. Changed language throughout to indicate that the null model of the system is used to test the null hypothesis.

The model used to test the null hypothesis ($H_0$) uses a modified formulation of the fitness functions.
The null hypothesis is that the need to identify messages from members of the same species \textit{does not} play a role in niche differentiation.
Therefore receivers are assumed to be able to know \textit{a priori} which messages come from senders of their own species and no fitness is assigned for the task of species identification in this null model.

In the null version of the model, this results in the receivers only processing messages from members of their own species and ignoring messages from the other species.
Senders and receivers are evaluated with the fitness function in \autoref{eq:total} with $e_s=0$.

\newcommand{\nruns}[0]{{$20$}}

\section{Results}
\label{sec:results}

%TODO RESP:  REV: "the signals are never overlapping (so there is no need to differentiate the frequencies)" || TODO DK: make task more clear. Show that the signals overlap in the first few generations, over time they separate. Talk about “OVERLAP” earlier on in the paper. Use the word “OVERLAP” in discussion of biological niche. Here is a FFWD simulation.
\begin{figure*}
    \centering
    	\includegraphics[width=\textwidth, trim=0 0 50 0, clip]{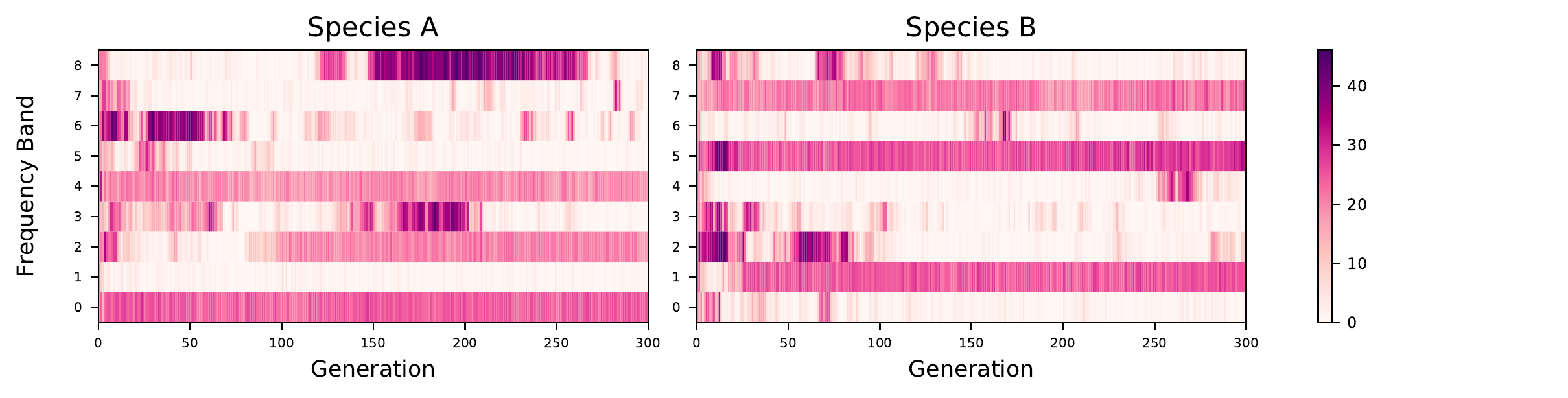}
    \caption[The use of the 9 frequency bands by the population of senders in the example simulation.]{The use of the 9 frequency bands by the population of senders in the example simulation of $H_1$.
    The graph shows the changing use of the frequency bands over 300 generations.
    In the first generations, both species's signals are spread across the 9 bands such that the signals from the two species overlap.
    These signals converge rapidly to a smaller subset of the available bands.
    By about the 50th generation, there is little overlap between the two species --- \textit{A} uses mainly bands 0, 4, and 6 while \textit{B}'s signals are concentrated on 1, 5, and 7 --- though there is some use of bands 2 and 3 by both species.
    In this example, \textit{Species B} uses band 2 heavily but intermittently until just before the 100th generation, when it ceases almost all activity on the channel and \textit{Species A} begins to make consistent use of it for the remainder of the simulation.
    By the generation 300, both species have converged to the near-exclusive use of 3 channels: 0, 2, and 4 for \textit{Species A} and 1, 5, and 7 for \textit{Species B}.}\label{fig:spectra}
\end{figure*}

We ran simulations of our ecosystem with senders and receivers for two species.
Each population consisted of 50 individuals and the simulation was run for 300 generations.
The results discussed here are averages and standard deviations from \nruns{} independent simulations.
Additionally, the results from a representative example simulation are highlighted in figures and throughout this section in order to discuss specific features of an individual simulation.

For each simulation, we generated spectrograms that mirror the type of chart that is often presented in studies of soundscapes~\citep{Krause1987,Pijanowski2011a}, except that the $x$-axis of these plots represents generations instead of real-time auditory signals.
These diagrams, such as the one seen in \autoref{fig:spectra}, show how the two species's use of the frequency bands shifts from generation to generation.
The initial populations's encoded messages are randomly distributed across the 9 frequency bands, but the signals converge over the course of the first 50 to 100 generations into a subset of bands used primarily by one species.
In this example, after an initial series of about 100 generations, both species show consistent use of 3 bands --- 0, 2, and 4 for \textit{Species A} and 1, 5, and 7 for \textit{Species B} --- for the remainder of the simulation.
\textit{Species A} develops and then eventually scales down the use of band 3 and band 8, but \textit{Species B}'s use of 1, 5, and 7 remains remarkably stable through most of the latter 200 generations.

\begin{figure*}
    \centering
    \begin{subfigure}[b]{\linewidth}
    	\includegraphics[width=\linewidth]{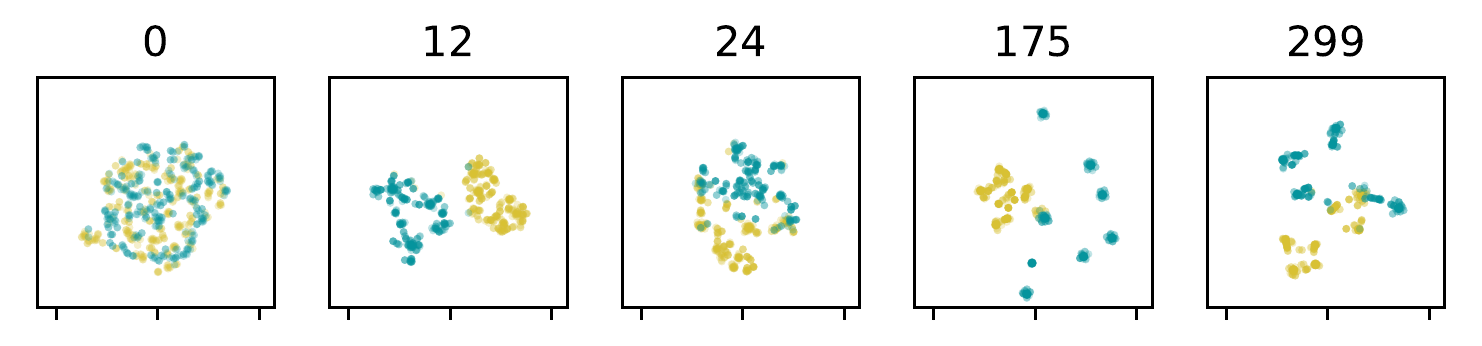}
        \caption{Cluster diagrams of messages in selected generations, mapped to 2 dimensions using t-SNE. Each point is a message generated by a sender in a single generation (labeled above the plot) of the simulation. The selected generations are marked on the silhouette score plot below with pink circles. The different colours represent messages originating from members of the two different species. The messages in generation 0 are scattered randomly from both species as the initial neural network connections for the senders are randomly generated. The messages rapidly converge to two clusters by generation 12. However, these clusters are still evenly spaced internally, as the initial selection pressure is mainly to differentiate messages between the two species. In later plots, for example in those from generations 175 and 299, smaller clusters form within the messages from a single species as the senders from each species converge on representations for particular bits and messages. This clustering drives the increasing bit and total scores in \autoref{fig:score}.}
        \label{fig:cluster}
    \end{subfigure} \\
    \begin{subfigure}[b]{\linewidth}
        \includegraphics[width=\linewidth]{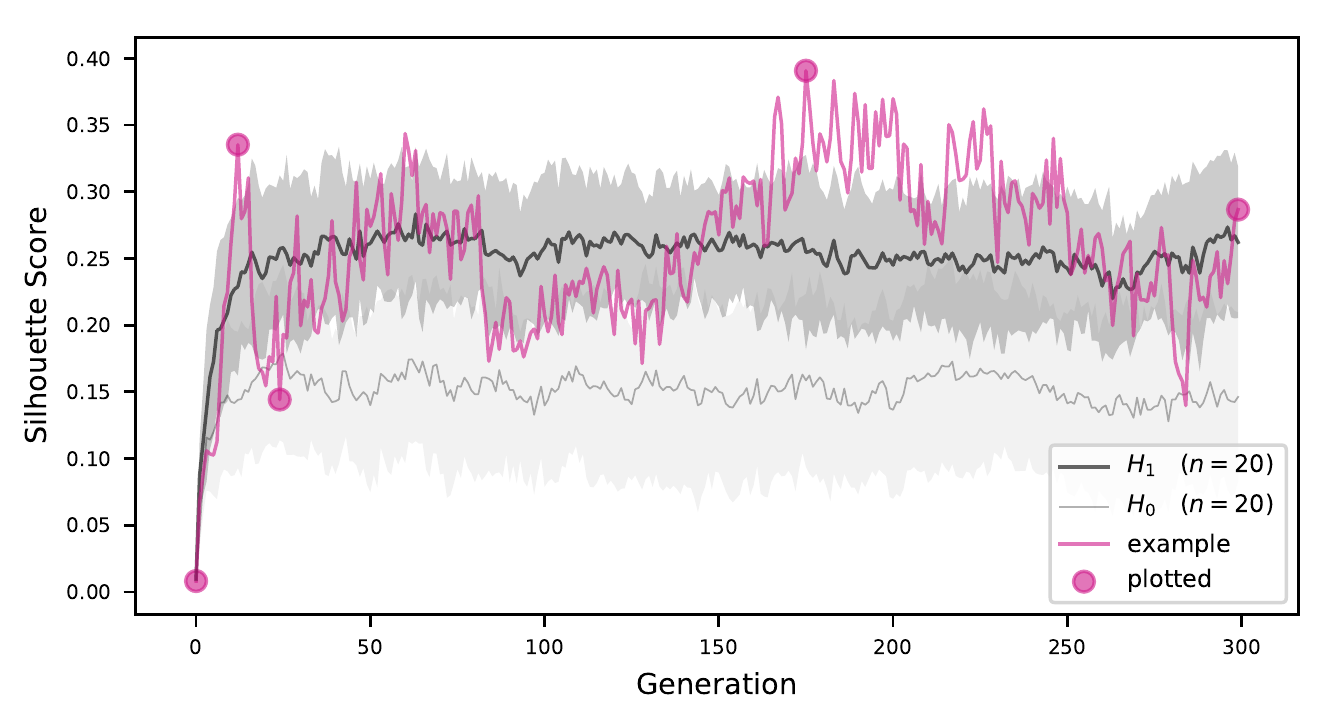}
        \caption[Silhouette score of the encoded messages, grouped by species, over the course of 300 generations.]{Silhouette score of the encoded messages, grouped by species, over the course of 300 generations. Scores reflect the validity of the message clusters when grouped by species, averaged over \nruns{} runs of the simulation, and plotted with the standard deviation in the background. An example of an individual run is also plotted (pink) and the generations of that run that are plotted in the cluster diagram above are noted. The difference between the alternative hypothesis ($H_1$) and the null hypothesis ($H_0$) is significant ($P<0.01$) after generation 4. The average P-value after generation 4 is $15\times10^{-5}$.}
        \label{fig:silhouette}
    \end{subfigure}
    \caption{Cluster validity scores over \nruns{} runs of the simulation. Message clusters are shown above for selected generations of an example run.}\label{fig:clustering_metrics}
\end{figure*}

The spectral plots provide a useful visual representation of the divergent signals, but the actual level of separation can be quantified further and visualized in another manner.
\autoref{fig:cluster} shows a mapping of the high-dimensional messages to two-dimensions using \gls{tsne}, plotted for particular generations of interest.
The encoded messages generated by senders from the two species can be seen to rapidly separate into clusters from an initial state of near-total overlap.
This can be further examined in the plot below the cluster maps (\autoref{fig:silhouette}) which shows the silhouette score for the clusters over the course of generations.
The silhouette score is used in the evaluation of clustering algorithms and is a measure of the density of clusters~\citep{Rousseeuw1987}, where a score of $0$ indicates overlapping data and a score of $1$ indicates dense and well-separated clusters.
The rapid rise of the silhouette score here indicates the splitting of the spectrum audio spectrum between the senders in relatively few generations.

The plot shows the average and standard deviations of the silhouette scores from the series of \nruns{} trials of $H_1$ (dark grey) alongside the silhouette score from the specific run from which the clusters in the plot above were derived (pink).
In addition, it shows the average and standard deviation of silhouette scores from \nruns{} trials of the null model $H_0$ (light grey).
A test of the hypotheses using Welch's t-test --- because the variance of the samples cannot be assumed to be equal --- reveals that the difference between the two models is significant after generation 4 ($P < 0.01$), with an average P-value of $15\times10^{-5}$ for latter 295 generations.

While the null model does produce a level of clustering of the species' messages, this is to be expected as a result of the selection of frequency bands on which to communicate.
However, in the null model, this selection is not competitively driven by the presence of the other species.
In $H_1$, the receivers of the two species drive their senders towards diverging frequency bands as their fitness increases with their ability to identify messages from their own species and reject those from the other.

%TODO RESP: REV: "Figure 3: by visual inspection, the t-SNE diagram of generation 24 seems better than the t-SNE diagram of generation 12, but the silhouette score of generation 12 is higher than the silhouette score of generation 24. Could the authors add some explanation or some comments about this issue?" Not sure why that is. || TODO DK: It’s important to remember that the t-SNE is a 2D representation of a multidimensional system so it is possible that in other dimensions the data separate better. Sebastian? Note in figure 3 about TSNEs and the visual separation ________ Silhouette score is calculated over N-dimensional data, while t-sne is a visualization in two dimensions so, for example, 12 and 14 look different in silhouette and t-sne...both are useful representations. Could attempt different p??? values for t-sne diagrams.

\begin{figure}
    \centering
	\includegraphics[width=\linewidth, trim=0 0 0 0, clip]{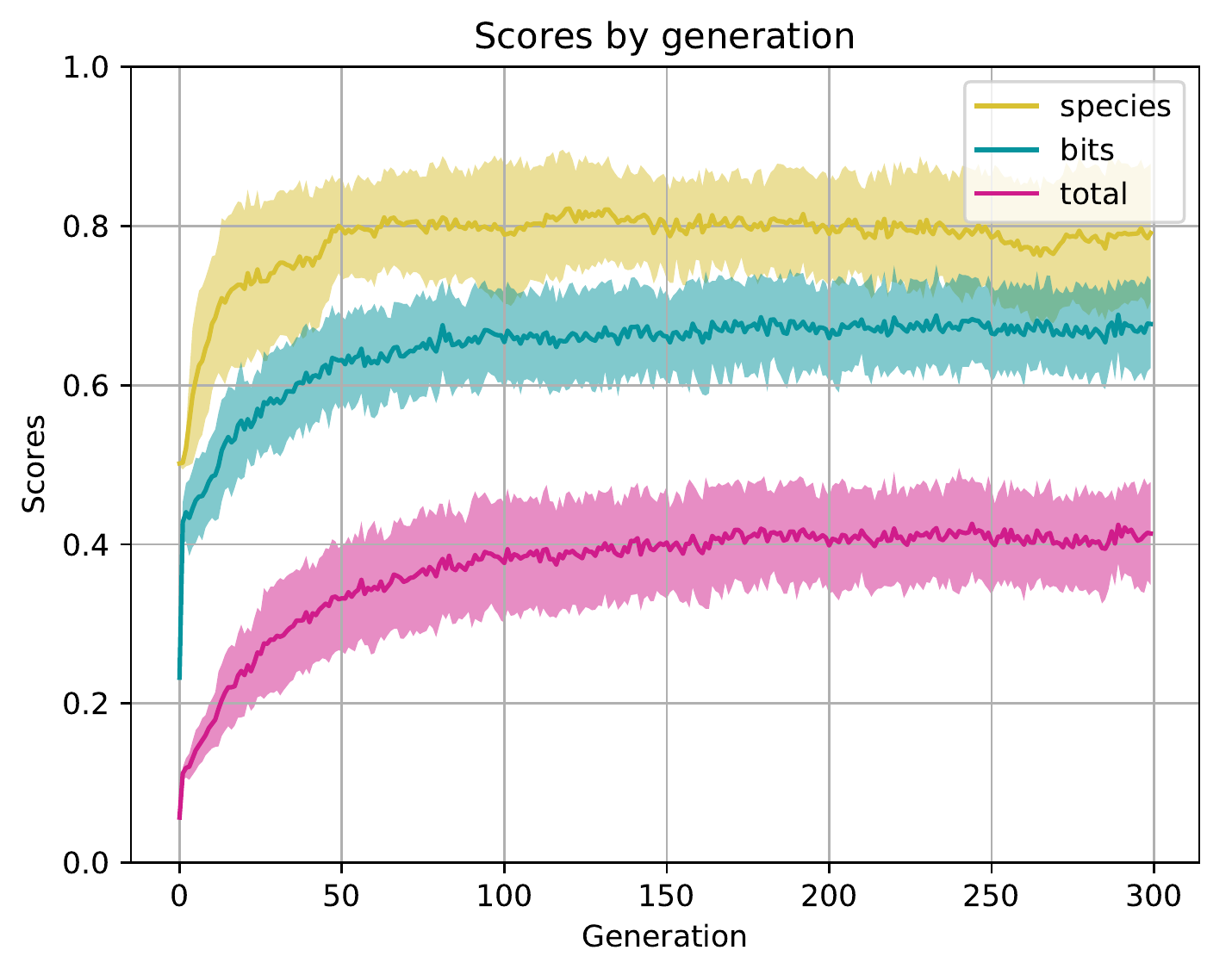}
    \caption{Performance of a species, showing the receivers' ability identify the species (yellow), and the rate at which sender-receiver pairs were able to correctly identify individual message bits (blue) and the entire message (pink).
    }
    \label{fig:score}
\end{figure}

%TODO: Is this actually relevant to the argument?
We also examined the actual performance of the species with regard to their ability to recognize and decode messages from their conspecifics.
\autoref{fig:score} shows the scores of the senders and receivers from a species over the course of 300 generations.
On average, the proportion of messages that are correctly identified as being from members of the same or other species (red) rises sharply in the first generations before steadying near 80\%.
The proportions of bits that are correctly decoded and messages that are fully decoded correctly are slower to rise, but continue to do so throughout most of the evolutionary process.

\section{Discussion}

The results presented in the previous section demonstrate that it is possible to drive spectral differentiation in the acoustic signature of an agent through an impetus to communicate with other members of the same species.
An analysis of the distance between intraspecies messages and interspecies messages shows a significant difference between the test of the alternative hypothesis ($H_1$) and the null hypothesis ($H_0$), as seen in \autoref{fig:silhouette}.
Moreover, \autoref{fig:spectra} provides a visual reference for the division of the spectrum in a selected simulation of the alternative hypothesis ($H_1$).
The spectrum has been split between the two species after the first 100 generations, such that \textit{Species A} primarily makes use of bands 0, 2, and 4 while \textit{Species B} relies on bands 1, 5, and 7.
It is interesting to note that, in the first 100 generations, band 2 is used mainly by \textit{Species B}, however this changes around generation 90 as \textit{Species A} begins to use the band regularly.
Once \textit{Species A} establishes regular use of the band, \textit{Species B} never returns to it with any stability for the remainder of the simulation.

In models of the null hypothesis ($H_0$), the two species occasionally achieve a level of differentiation of their messages, however this occurs only by chance.
In both models, species tend to converge to the primary use of roughly 3 of the 9 available channels for communication.
Three channels is the fewest that can be used to encode the three-bit message and it is often the easiest solution for the evolving neural networks to find.
However, in the null case, the channel selection is not driven by competition between the species, only by cooperation within a species.
This lack of competition often leads to overlapping channel selections, which in turn, is responsible for the lower silhouette scores for the null models (\autoref{fig:silhouette}).

While these results cannot be taken as confirmation of the proposed mechanism of the \gls{anh}, they demonstrate that the mechanism is plausible.
The drive to produce signals that are identifiable and understandable to members of one's own species within the finite resource that is a soundscape results in the formation of acoustic niches for vocalizing species.

This study also demonstrates the efficacy of a highly simplified model in demonstrating the plausibility of a particular mechanism for the formation of patterns within a soundscape.
It compliments the work of \citep{Eldridge2018}, which explores the way that common acoustic indices respond to changing populations and signals, and presents another application for a synthetic acoustic ecology.
%TODO RESP: Talk about differences from eldridge/keifer here
Together with other types of computational studies of soundscapes~\citep{Eldridge2018}, this paper lays the foundation for a method of rapidly interrogating evolutionary acoustic processes.
In addition to providing insight into ecological studies, research in this area can also be used to inform the development and analysis of evolutionary acoustic agents live "in the wild" and interact with biological ecosystems.

\section{Conclusion}

Though the experiment presented here is based on a highly simplified model of a physical ecosystem, it demonstrates that it is possible to rapidly and repeatedly test some of the basic principles of soundscape ecology.
As predicted, the experiment was able to demonstrate the important role of intraspecies communication in the partitioning of the acoustic resources of an ecosystem.

This has important implications for the development of hardware-based \gls{alife} agents for the production of sound in a physical, hybrid ecosystem.
It suggests that, if one of the goals of that agent is to identify a niche for itself in the soundscape, it is important to co-evolve the auditory production with auditory perception to drive the vocalizations into an empty portion of the spectrum.

In a broader sense, this experiment sets out the foundation for a method of testing ideas for hardware-based agents in software simulations to understand the possible dynamics once they are released in the field.
It grounds the inquiry into a complex phenomenon with a concrete example that solidly demonstrates the theoretical basis for a physical experiment through repetition and statistical analysis on a scale that is difficult to achieve in the field.
And it demonstrates the feasibility of a key theory in soundscape ecology.

\section{Acknowledgements}

Thanks to the REAL Lab, especially Kasper Støy, Rosemary Lee, Djordje Grbic, Miguel Gonzalez Duque, Mads Johansen Lassen, and Niels Justeen for their suggestions.

\footnotesize
\bibliographystyle{apalike}
\bibliography{library} % replace by the name of your .bib file
%\bibliography{/Users/davk/bibtex/library} % replace by the name of your .bib file
%\printbibliography%[title=Current Bibliography]

\end{document}